\documentclass[sn-mathphys]{sn-jnl}

\jyear{2021}%

\theoremstyle{thmstyleone}%
\theoremstyle{thmstyletwo}%
\usepackage{adjustbox}
\theoremstyle{thmstylethree}%
\usepackage{natbib}
\usepackage{tipa}
\usepackage{algorithm,algorithmic}
\usepackage{subcaption}
\usepackage{arydshln}
\usepackage{enumerate}
\begin{document}

\title[The Development of a Labelled te reo M\={a}ori-English Bilingual Corpus]{The Development of a Labelled te reo M\={a}ori-English Bilingual Database for Language Technology}


\author*[1]{\fnm{Jesin} \sur{James}}\email{jesin.james@auckland.ac.nz}

\author*[1]{\fnm{Isabella} \sur{Shields (Ng\=ati Porou)}}\email{ishi836@aucklanduni.ac.nz}

\author[2]{\fnm{Vithya} \sur{Yogarajan}}

\author[3]{\fnm{Peter} \sur{J. Keegan (Waikato-Maniapoto, Ng\=ati Porou)}}

\author[1]{\fnm{Catherine} \sur{Watson}}

\author[4]{\fnm{Peter-Lucas} \sur{Jones (Ngāti Kahu, Te Rārawa, Ngāi Takoto,Te Aupōuri)}}

\author[4]{\fnm{Keoni} \sur{Mahelona}}

\affil*[1]{\orgdiv{Department of Electrical, Computer, and Software Engineering}, \orgname{The University of Auckland},  \country{New Zealand}}

\affil[2]{\orgdiv{Strong AI Lab}, \orgname{The University of Auckland},  \country{New Zealand}}

\affil[3]{\orgdiv{Te Puna Wānanga }, \orgname{The University of Auckland},  \country{New Zealand}}

\affil[4]{\orgdiv{Te Hiku Media},  \country{New Zealand}}

\abstract{ Te reo M\={a}ori (referred to as M\={a}ori), New Zealand's indigenous language, is under-resourced in language technology. M\={a}ori speakers are bilingual, where M\={a}ori is code-switched with English. Unfortunately, there are minimal resources available for Māori language technology, language detection and code-switch detection between M\={a}ori-English pair. Both English and M\={a}ori use Roman-derived orthography making rule-based systems for detecting language and code-switching restrictive. Most M\={a}ori language detection is done manually by language experts. This research builds a Māori-English bilingual database of 66,016,807 words with word-level language annotation. The New Zealand Parliament Hansard debates reports were used to build the database. The language labels are assigned using language-specific rules and expert manual annotations. Words with the same spelling, but different meanings, exist for Māori and English. These words could not be categorised as Māori or English based on word-level language rules. Hence, manual annotations were necessary. An analysis reporting the various aspects of the database such as metadata, year-wise analysis, frequently occurring words, sentence length and N-grams is also reported. The database developed here is a valuable tool for future language and speech technology development for Aotearoa New Zealand. The methodology followed to label the database can also be followed by other low-resourced language pairs.}


\maketitle

\section{Introduction}\label{sec1}
The UNESCO International Year of Indigenous Languages (IYIL2019), which aimed to protect and promote the use of indigenous languages around the world, took place in 2019. The Language Technologies for All Conference (LT4All) was organised in the framework of IYIL2019, with the following overarching principle:

\begin{quote}
 ``Everyone should have the possibility to get access to language technologies in his/her native languages, including indigenous languages.'' 
 \end{quote}
 
LT4All highlighted the importance of developing language technologies, especially since technologically under-resourced languages stand to be used less and less. Furthermore, the needs of an indigenous language are particular to the language, and the needs of its speakers. 

Te reo Māori (meaning the Māori language, hereafter referred to as M\={a}ori), the indigenous language of Aotearoa\footnote{Aotearoa is most commonly used M\={a}ori name for New Zealand.} has had extensive language contact with English resulting in detrimental effects on the usage of M\={a}ori \cite{Eliasson_1989_TRM}. In response to this decline, extensive M\={a}ori language revitalisation efforts since the 1980s have uplifted the use of the language and created awareness among people about the importance of M\={a}ori revitalisation. In New Zealand, M\={a}ori-language revitalisation is viewed as a means to re-establish ties with māoritanga (Māori culture, beliefs, and practices) and, for some, with the whenua (the land). It is imperative that M\={a}ori language resources and technology are co-designed and co-developed with M\={a}ori\footnote{Both the people and the language have the same name - M\={a}ori.} people, as it is the speakers of the language who best understand the sentiment behind revitalising the language and thereby contribute to resources that best capture the language. There is also consensus in the research community regarding the need to decolonise speech and language technologies for indigenous languages \cite{bird-2020-decolonising}. The paper also strongly suggests the need for co-designing language technology by treating its development as a cross-cultural endeavour, that identifies the community's goals, with a sustainable use model in place and broadening technology evaluation. Te Hiku Media\footnote{https://tehiku.nz/te-hiku-tech/papa-reo/}, a charitable media organisation that belongs to and serves Māori, has pioneered significant advances in speech and language technology development for M\={a}ori. The research presented in this paper has been developed in consultation with Te Hiku Media as an ongoing effort to co-design M\={a}ori language resources.

Initial speech synthesis, speech recognition systems, and pronunciation tools are under development for M\={a}ori~\cite{James_2020_TRMTTS, Tehiku_Speech_Recognition, mpai_2017}. The main challenge with M\={a}ori speech and language data is annotation. Generally, speakers of M\={a}ori are bilingual and usually speak English fluently. Code-switching, a phenomenon that explains the fluid alternation between two or more languages~\cite{gumperz1982discourse}, of Māori and English, takes place. The detection of these instances of code-switching is a prerequisite to annotating language and speech data. As English and M\={a}ori use the Roman orthography and an adapted Roman orthography, respectively, many annotations are done manually in the absence of automatic language detection systems. This makes the process time-consuming, thereby slowing down research and technology development. Such language detection is also essential for real-time speech recognition and synthesis systems. M\={a}ori is under-resourced in language technology resources; hence, the amount of publicly available M\={a}ori-English bilingual data is also limited. This research focuses on developing a fully annotated bilingual language corpus of te reo M\={a}ori and English with word-level annotations of the language. 

All language data collection for the development of automated technology in this research was done along with the M\={a}ori community, seeking feedback at every step and ensuring data sovereignty and trust \cite{data_guardian}. All the data collected and analysed in this study have been shared with us by M\={a}ori on trust. Hence, we shall remain guardians of the data we have in hand\footnote{M\={a}ori-world view regards data as taonga (unique resource). The researchers are the kaitiaki (guardians) of the data.}. All data developed or collected as part of this research is bound by the Kaitiakitanga license \cite{kaitiakitanga_license}, which recognises the importance of researchers being guardians of the data they collect. This ensures that the positive benefits of the data collected from Māori reach back to the community from which the data was collected. This further stems from Māori data sovereignty \cite{Data_sov}, which is being discussed and prioritised in Māori technology development and data collection.

The contributions of this research are:
\begin{enumerate}
\item The development of a gold standard language-annotated Māori-English bilingual database.
\item A detailed description of the database development process to support future research in under-resourced languages.
\item Analysis of the gold standard language-annotated bilingual database.
\end{enumerate}
 
The remaining sections of the paper consist of Section \ref{section:trm}, where the details of te reo Māori and the difference and similarities with English are presented. This is followed by Section \ref{section:hansard_db_dev}, which describes the development of the gold standard language-annotated Māori-English bilingual database. Section \ref{section:db_analysis} reports the analysis of the gold standard database. This is followed by Section \ref{section:discussion} where the results of the database analysis are discussed in detail, followed by Section \ref{section:conclusion}, which concludes the paper.

\section{Background and Review} \label{section:trm}

This section provides a brief overview of the M\={a}ori language and other resources available for language technology development in Māori.

\subsection{Te reo M\={a}ori (The M\={a}ori Language)} 
M\={a}ori is one of New Zealand's official languages (along with New Zealand Sign Language) and is New Zealand's only indigenous language. M\={a}ori is spoken by 4.5\% of the country's total population of 5 million \cite{NZ_stats}.  

Some language-based differences between te reo M\={a}ori and English are summarised in Table \ref{table:phonotactic_difference}. M\={a}ori phonology consists of 5 short vowels  /i e a o u/ and 10 consonants /p t k m n \textipa{N} f r w h/  \cite{Bauer_maori_1993}. In modern orthography, long vowels are denoted with a macron (e.g. \textit{ā}). In older texts, they are sometimes expressed as double vowels (e.g. aa), with an umlaut (e.g. ä), or ignored completely and represented the same way as a short vowel (Row 1, Table \ref{table:phonotactic_difference}). In addition, there is some regional variation in the way words are pronounced that can also be reflected in orthography (e.g. Aoraki /aoraki/ also appears as Aorangi /aora\textipa{N}i/\footnote{Aorangi/Aoraki is the name of a mountain in New Zealand.}). M\={a}ori orthography is highly phonemic; i.e., there is usually a one-to-one mapping of a phoneme to a grapheme, with some exceptions such as the two digraphs; \textless wh\textgreater, which is /f/, and \textless ng\textgreater which is /\textipa{N}/. This contrasts with English, which has a highly non-phonemic orthography (Row 4, Table \ref{table:phonotactic_difference}). 

English syllable structure allows for consonant clusters in both the onset and coda (end) of a syllable (Row 2, Table \ref{table:phonotactic_difference}). The syllable structure of M\={a}ori consists of a nucleus, which may be occupied by a vowel (or a diphthong), and an optional onset (syllable start) occupied by a single consonant. Hence, consonant clusters are not present in M\={a}ori \cite{harlow2007maori}. While M\={a}ori has several loan words from English, these have been adapted to match the available phonemes, graphemes, and phonotactic constraints of M\={a}ori. E.g., the word \textit{milk} (/\textipa{m\textsci lk}/, a single syllable) in English has been borrowed and adapted to \textit{miraka} (/\textipa{mi.ra.ka}/, three syllables) in M\={a}ori.

Sentences in M\={a}ori are described as being composed of phrases (Row 3, Table \ref{table:phonotactic_difference}). These phrases can have both syntactic and phonological significance. According to \cite{Biggs_1969}, a grammatical phrase in M\={a}ori consists of a nucleus (generally speaking, the lexical content of the phrase), a proposed periphery (a particle indicating, for example, a tense), and a postposed periphery (another particle). Depending on the type of particle in the proposed periphery, the phrase is designed either as a verb phrase or as a nominal phrase. The typical sentence structure of M\={a}ori is verb-subject-object. English, for comparison, has a subject-verb-object structure.

Some similar sounds in Māori and English are not always represented by the same orthographic character, despite both languages using a similar Roman-derived script. For example, Māori /k/ appears as \textless k\textgreater  while English /k/ appears as several different graphemes (\textless k c ck ch ...\textgreater). The graphs \textless b c d f g j l q s v x y z\textgreater   are permitted in English but not in Māori (Row 5, Table \ref{table:phonotactic_difference}).

 \begin{table}[!t] 
\caption{Some language-based differences between Māori and English}
\centering
\begin{tabular}{p{0.25cm}|p{6cm}|p{4cm}}
   \hline
   No.  & Differences & Example \\ \hline
1    & Macrons appear in Māori vowels but not usually in English  & Māori \textit{pōkē} vs. English \textit{poke}  \\ 
2    & Māori follows (C)V syllable structure, English (C)V(C) (with consonant clusters permitted) & English word \textit{ramp} not possible in Māori \\
3    & Different phrase structures & English: \textit{Peter ate oranges.}, Maori: \textit{Ka kai a Pita i nga okana.}; where \textit{kai\footnote{Ka kai = eat}} is the verb, \textit{Pita\footnote{Pita = Peter}} is the subject, \textit{okana\footnote{okana = orange}} is the object. \\
4    & Different phonemic-orthographic mappings                wh, ng phonemic-orthographic mappings & \textless wh\textgreater in Māori usually corresponds to /f/, but to /w/ or /h/ in English. \\
5    & Graphs \textless b c d f g j l q s v x y z\textgreater   not permitted in Māori  & Word \textit{fare} does not exist in Māori\\

    \hline 

\end{tabular}
\vspace{-1em}
\label{table:phonotactic_difference}
\end{table}

\subsection{Language technology development resources in Māori}

Māori language technology development has been active in recent years; however, major technology development is restricted by limited resource availability. Even though many sources of Māori written and spoken data are available as archival collections, they have not been processed and made into a usable form. This subsection provides an overview of the resources developed so far for te reo Māori. These resources are being extensively used for linguistic analysis. The use of these resources in language technology development is gaining momentum, with the need for more resources being acknowledged. When we comment on the language labels available for these databases, we refer to word-level labels indicating which language the word is.

The MAONZE corpus contains Māori speech and its corresponding text transcriptions, capturing the changes to the language over the years. The MAONZE corpus was collected to understand how the pronunciation of Māori had changed over time. Its motivation was sociolinguistic in nature \cite{king2011maonze}. However, in consultation with the community, it also became the exemplar in a Māori pronunciation tool \cite{mpai_2017}. The MAONZE project\footnote{\url{https://maonze.blogs.auckland.ac.nz/}} has contributors from across Aotearoa New Zealand. The MAONZE database contains both Māori and English. The MAONZE data was recorded either in English or in Māori, which was clearly labelled. However, if there was code-switching in either language, this was ignored. 

The Reo Māori Twitter (RMT) Corpus \cite{keegan2015RMT, trye2022harnessing}, a corpus of Māori-language tweets, and the Māori Loanwords on Twitter (MLT) corpus \cite{trye2019maori} are Māori databases that contain informal tweets data.   These corpora are used extensively for linguistic analysis. Automatic approaches using language-specific rules are used to label the language of words in the RMT corpus, but the labels have not been hand-checked. For the MLT corpus, the data (tweets) were manually labelled as `relevant' or `irrelevant' by the researchers.

The New Zealand Ministry of Education sponsored the Māori Niupepa (Newspaper) Collection \cite{Niupepa_data} consisting of historic Māori newspapers. It is based on the microfiche collection produced by the Alexander Turnbull Library. Other universities in New Zealand also sponsor the initiative. These collections are in Māori, with English translation available in some cases. However, language labelling is not available.

The Ngā mahi corpus \cite{James_2020_TRMTTS} consisting of speech and text from a single Māori speaker was developed for initially for text-to-speech synthesis development. This corpus has been used for Māori text-to-speech synthesis development. The database is purely in Māori but is not publicly available. 

He Pātaka Kupu Ture or the Legal Māori Corpus \cite{LMC_data} is a collection of Māori and English sentences used in a legal context collected by the Victoria University of Wellington researchers. This corpus provides a rich resource of Māori written data and has been used for linguistic analysis. This database contains both Māori and English, with no language labelling done.

Te Hiku Media is an organisation committed to revitalising tikanga (Māori customs and traditional values) and te reo Māori and is developing the Papa Reo project \cite{Tehiku_media1} - a language platform for a multilingual Aotearoa. Te Hiku Media has pioneered multiple speech and language technology developments for te reo Māori. A corpus of te reo Māori derived from the New Zealand Hansard reports was developed by Te Hiku media \cite{Tehiku_research}. Automatic approaches using language-based rules are used to label the language of words in the corpus. Other than the language labels (Māori or English), two more labels - Ambiguous (words having the same spelling in Māori and English) and Unclear (the automatic approaches cannot identify the language of the word) are also present in the database. These labels have not been hand-corrected.

The above-described resources are the only open-source or available-upon-request Māori text databases available. Although these resources contain extensive Māori usage, the language labels are not available, or they are not manually hand-checked by an expert due to Māori and English sharing the same Roman script, hand-checking and correction of labels in essential. Understanding the need for a language-labelled Māori-English corpus for language technology development, we describe the collection, labelling, hand-checking and analysis of the Hansard corpus in the following sections.

\section{Development of Hansard Māori-English Labelled database}
\label{section:hansard_db_dev}

To conduct Māori-English language detection and code-switching detection analyses, it is essential to have a large-scale database that has the two languages along with the language of each word labelled. Such a database did not exist at the beginning of this research. Hence, we collected and labelled a database. This section presents the development and manual labelling of the `gold standard' Hansard database. 

\subsection{Text collection and pre-processing} \label{hansard_db}

 \begin{figure*}
    \centering
  \includegraphics[scale=0.35]{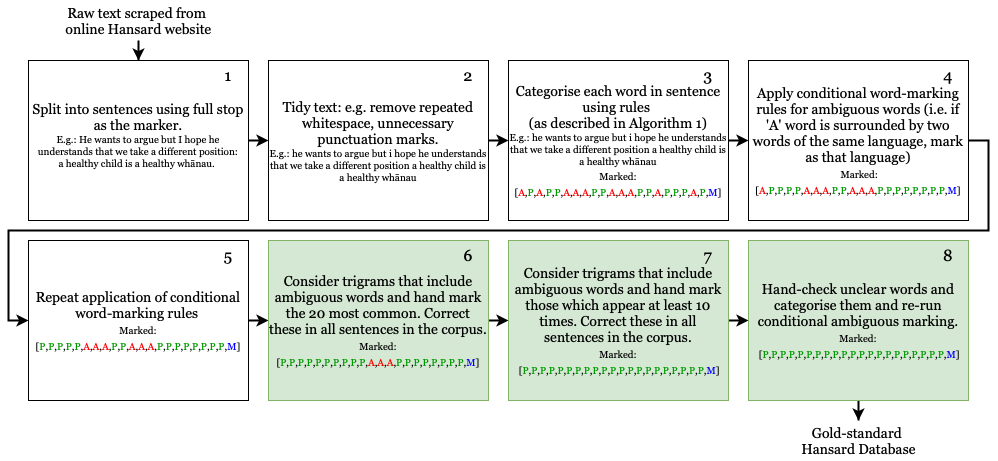}
    \caption{Steps followed to obtain the Gold standard Hansard database. The blocks marked in green required manual labelling.}
    \label{fig:label_block}
\end{figure*}

 The Hansard database consists of text transcripts from the official reports of debates in the New Zealand House of Representatives (New Zealand Parliament). It is, therefore, a transcription of spoken language in a mostly formal and political context. While several different languages are present in these, most words are English or M\={a}ori. Debates from 2003 are available as HTML-formatted web pages on the New Zealand Parliament's website\footnote{\url{https://www.parliament.nz/en/pb/hansard-debates/rhr/}}.  
 
  \begin{table}[!t]
\caption{Details of initial labels used in the Hansard database.} 
\centering
\begin{tabular}{p{2cm}|p{9cm}}
   \hline
    Label    & Details \\
    \hline 
        \textit{Number (N)} &  Numeric entries \\
      \textit{Symbol (S)} &  Punctuation symbols \\
      \textit{Ambiguous (A)}  &  Ambiguous words that have the same spelling in both languages  \\
    &Examples: \textsf{home, hope, kia, rite, kite} \\
    \textit{M\={a}ori (M)} & M\={a}ori words \\
    \textit{English (P)} & Pākehā\footnote{The Māori word \textit{pākehā} can refer to someone or something being of NZ European or foreign descent. Here, we designate English words as \textit{pākehā} and give the label `P'.} English words  \\
    \textit{Unclear (U)} &  None of the above \\
 
    \hline
\end{tabular}
\vspace{-1em}
\label{table:labels_for_text}
\end{table}

 \begin{algorithm}[t]
\caption{Algorithm used for rule-based labelling of the Hansard database.} 
	\label{alg:rules}
	\begin{algorithmic}[1]
		\STATE\textbf{Input:} A sentence 
		\STATE\textbf{Output:} Labels at word level
		\FOR {each sentence i}
		    \FOR {each word in sentence i}
    	        \IF{word is number }
    	            \STATE label is `N'
    	        \ELSIF{word is punctuation}
    	            \STATE label is `S'
    	        \ELSIF{word is in ambiguous words list }
    	            \STATE label is `A'
    	        \ELSIF{word is in M\={a}ori Dictionary or use macrons [\={a},\={e},ī,\={o},\={u}]}
    	            \STATE label is `M'
    	        \ELSIF{word is in English Dictionary or illegal characters for M\={a}ori, \\ i.e. [b,c,d,f,j,l,q,s,v,x,y,z]}
    	            \STATE label is `P'   
    	        \ELSE
    	            \STATE label is `U'
    	        \ENDIF     
            \ENDFOR
 		\ENDFOR

	\end{algorithmic}
	\end{algorithm}
 
 The Hansard debate transcripts were scraped from the New Zealand Parliament's website to build this corpus. This process was adapted from that used by Te Hiku Media in their scraping of the debates for analysis of the increasing presence of M\={a}ori words\footnote{\url{https://github.com/TeHikuMedia/nga-kupu}, \newline \url{https://github.com/TeHikuMedia/nga-tautohetohe-reo}}.

 Figure \ref{fig:label_block} illustrates the steps followed to obtain the gold-standard Hansard database from the scraped text. First, the scraped text was split into sentences using the full stop as the marker. Then the sentences were tidied up by removing repeated white spaces and unnecessary punctuation marks (Steps 1, 2 in Figure \ref{fig:label_block}).

 \subsection{Word-level language labelling} \label{sec_label}
 \subsubsection{Dictionary- and rule-based labelling}
 
 Each word in the database was categorised into six labels - `N', `S', `A', `M', `P' and `U', as detailed in Table \ref{table:labels_for_text}. The labels `N' and `S' capture digits and punctuation symbols, respectively, while the remaining labels capture words. Algorithm~\ref{alg:rules} provides an overview of the rules initially used to label each word of the Hansard database. The algorithm also indicates the order in which the categorisation of words into the different categories was done. Each word scraped was compared against existing dictionaries for M\={a}ori, English and Ambiguous list\footnote{The Māori and English dictionaries are a result of previous research by first, second and fourth authors of this paper, while the Ambiguous dictionary was extended from that developed by Te Hiku Media available as kupu\textunderscore rangirua.txt here \url{https://github.com/TeHikuMedia/nga-kupu/}} and assigned a label corresponding to the language (\textit{M\={a}ori} with the label `M' or \textit{English} with the label `P). The existing dictionaries for Māori words were modified so that alternate spellings of words without macrons were also marked `M'. This means, for example, the word \textit{māori} was recognised as `M' whether it was spelled \textit{māori}, \textit{maaori}, or \textit{maori}.
 
 \textit{Ambiguous} label (`A') was used for words that share the same spelling in the two languages, such as \textit{rite} (M\={a}ori: to be like, resemble; English: ceremony, ritual). Ambiguous words present a challenge for language detection, given they can appear as either M\={a}ori or English. \textit{Unclear} label (`U') was used for words that did not appear in any dictionary used or did not match phonotactic checks. These checks were based on the phonotactic differences between the two languages, framed as rules in Algorithm~\ref{alg:rules} lines 11 to 14. The presence of consonant clusters, aside from `ng' and `wh', resulted in the word being labelled `P'. This was also the case for any words which had non-Māori characters, such as `b'. However, this was not without issue. Māori words that had undergone English pluralisation, that is, where `s' is suffixed to a Māori word, are marked as `P'. For example, the Māori word \textit{iwi}, when pluralised to \textit{iwis}, is not marked `M' but rather `P'. The label `P' also captures some words that are neither Māori nor English. For example, in the following corpus sentence, the French word \textit{bonjour} has been automatically labelled `P' based on the presence of the characters `b', `n', and `j': 
 
 \begin{quote}
 ``Yes, bonjour to our friends who have a French background in the Pacific region.'' 
 \end{quote}
 
 Correct labelling of these other languages is outside of the scope of this corpus labelling task. However, these incorrectly labelled words comprise only a small portion of the overall database.

The \textit{Ambiguous} and \textit{Unclear} labels together contributed to 1.8\% of words in the Hansard database (A words: 1,129,045 [1.7\%]; U words: 57,946 [0.1\%]; A or U in a sentence: 664,851 [0.2\%]). These words require manual checking. We framed more rules derived by inspecting the Hansard database to reduce this mammoth task. This stage corresponds to the conditional marking stage in Figure \ref{fig:label_block} (Stage 4). This conditional marking considered the context of the \textit{ambiguous} word by looking at the marking of each word in a sentence. Based on our observation of the database, we made the condition that an \textit{ambiguous} word with a \textit{Māori} word preceding and following should be marked as \textit{Māori} (`M'). Conversely, an \textit{ambiguous} word with \textit{pākehā} words in its immediate vicinity should also be marked as \textit{pākehā} (`P'). In instances where the \textit{ambiguous} word occurred at the beginning of a sentence, the following two words had to share the same language marking (`M' or `P') to be changed. This conditional marking process was applied twice and successfully reduced \textit{ambiguous} labelled words to 1.5\% of the Hansard database (A words: 1,006,440 [1.5\%]). 

\subsubsection{Hand-labelling of ambiguous and unclear words}
Further steps to label the remaining \textit{Ambiguous} and \textit{Unclear} words relied heavily on hand correction completed by the authors. The first stage of hand correction was the collection and sorting of trigrams that included ambiguous words. As described in Figure \ref{fig:label_block} (Steps 6, 7), first, the 20 most common trigrams in the database that contain ambiguous words were identified, and the ambiguous words were manually marked as `M' or `P'. For example, the trigram \textit{I make a} is composed entirely of ambiguous words which, when combined in that order, are clearly English by visual inspection and were thus marked `P'. From the trigrams with ambiguous words that remained, those that appeared at least 10 times were identified, and the ambiguous words were manually marked as `M' or `P'. We also updated the ambiguous words list during this process. At this stage, the resulting database consists of 66,016,807 words, with 790,839 being Māori and the remaining being English, unclear (i.e. to be assigned to a dictionary), ambiguous (i.e. unable to be assigned on simple rules), a numerical value or a symbol. 

Following the hand-labelling of ambiguous words, we considered the unclear words. As previously mentioned, unclear words were defined when a word did not meet the criteria of any other label. All unclear words were labelled but could not be captured by the six label categories discussed above. Most notably, words from languages other than Māori or English were not effectively captured and were labelled `F'. The range of languages in the corpus reflects the multicultural nature of Aotearoa; there was a notable presence of words from Samoan, Fijian, Tongan and Niuean. It is out of the scope of this database development to label other languages in the database.

In addition, several unclear words were what we have determined to be dialectal or `alternative' spellings of words. For example, the word \textit{w'akarongo} occasionally appears instead of the usual \textit{whakarongo}. Cook Islands Māori words were also present in the Hansard data and considered separate from Māori. Some Cook Islands Māori words were likely recognised as Māori; however, these comprise a very small portion of the database. 

The remaining \textit{Unclear} words consist primarily of written errors from transcription/digitisation at the publication stage and were not rectified here. These, along with `S', `N', and the remaining `A' words, are excluded from the final labelled version of the gold standard database, which includes only words labelled `M' or `P'.

We regularly updated the Māori, English dictionaries and the Ambiguous words list during the manual labelling process. The updated dictionaries were used to re-assess each word considered \textit{Unclear}, followed by a repetition of the conditional ambiguous marking discussed previously (Stage 8 Figure \ref{fig:label_block}. The result of this stage was 842,680 Māori words (1.3\%), 63,050,059 English words (95.5\%), 949,240 (1.4\%) \textit{Ambiguous} words, and 10,833 \textit{Unclear} words. 

Labelling words in the database was a time-consuming task that took six months to complete. It was essential to get the database labelling accurate, as this database is the first completely labelled bilingual M\={a}ori-English text database. The accurately labelled Hansard database was used as the gold-standard database for our code-switching detection experiments. This will be made available for other researchers upon request bound by the Kaitiakitanga license and can be used to train and evaluate language models for detecting code-switch.

\subsection{Changes in Māori and English word count during database processing}
\begin{figure*}
    \centering
  \includegraphics[scale=0.5]{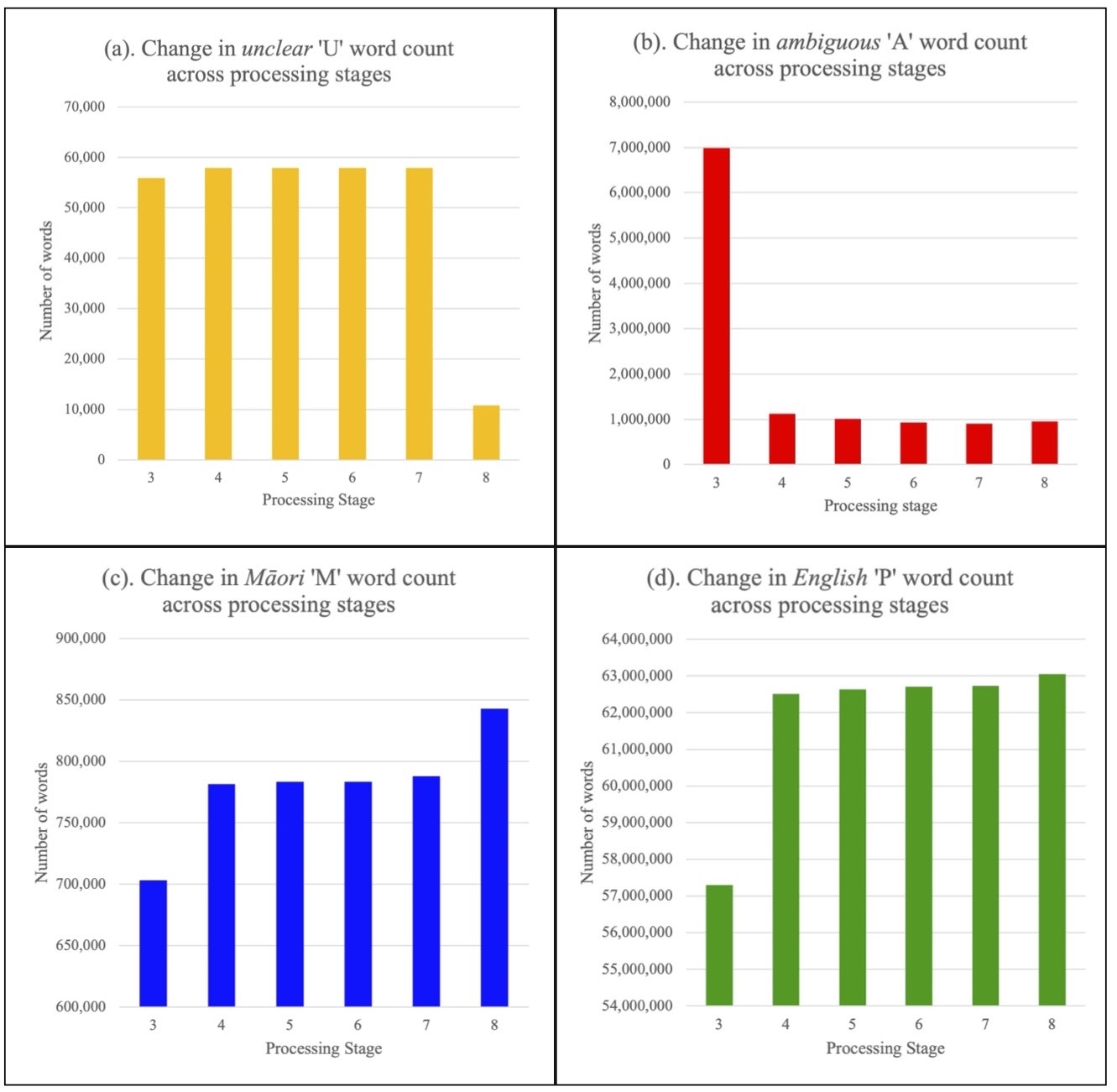}
    \caption{Changes in number of words across corpus processing stages: Subfigures (a), (b), (c), and (d) show the changes in \textit{unclear} words, \textit{ambiguous} words, \textit{Māori} words, and \textit{English} words, respectively.}
    \label{fig:word_count_changes}
\end{figure*}

With each database processing stage, there were changes in the overall word counts for each category. In Figure \ref{fig:word_count_changes}, we summarise the changes from stage to stage, focusing in particular on the categories of `M', `P', `A', and `U'. These are accompanied by the changes in the total word count for the database. The processing stages are marked with reference to the stage numbers in Figure \ref{fig:label_block}.

The overall goal of each of the stages outlined in Section \ref{sec_label} was to increase the number of Māori and English works marked as `M' and `P', respectively. Given the unbalanced nature of the corpus, the priority was to increase the number of Māori words. Different stages had different degrees of success, and improvements generally became more and more incremental. Each stage reflects the biased nature of the corpus, with significantly more English words appearing at each stage than Māori words. 

Stage 4, where ambiguous words `A’ are conditionally marked as `M’ and `P’, resulted in the largest overall change in markings, with 5,839,000 ambiguous words being categorised. Of that number, 88,000 were marked as Māori. Stage 5, which involves the re-application of the conditional marking process applied in Stage 4, yielded a smaller change, with 119,000 ambiguous words updated, of which 2,000 were Māori. 

Stages 6 and 7 both adjust `A’ words but are based on hand-marked trigrams. The focus of Stage 6 was the 20 most common trigrams containing an ambiguous word. This stage only impacted English words, with 77,000 words newly marked `P’. Stage 7, which affected 26,000 words, considered all trigrams with an ambiguous word that appeared at least 10 times. Of the 26,000 words adjusted, 4,000 words were marked as Māori. 

Between Stages 7 and 8, an error in the text processing of hyphenated words was identified and rectified. This resulted in an increase of 62,000 ambiguous words, 3,000 Māori words, 300,000 English words, and 5,000 Unclear words. The results of the final stage, Stage 8, combine the impacts of Unclear word labelling and a final conditional marking of ambiguous words. In total, 52,000 words were marked as Māori, and 21,000 words were marked as English. 

In total, the gold standard labelled  Hansard database consists of 842,680 `M'-labelled words, 63,050,059 `P'-labelled words, 949,240 `A'-labelled words, and 10,833 `U'-labelled words. Sentences with `A' and `U' words are excluded from the final bilingual database.

\subsection{Future Addition to Hansard Māori English Labelled Database }
The Hansard debates get added to the New Zealand Parliament website regularly. We aim to update the Hansard Māori English Labelled Database on a five-year basis. The newly available Hansard debates will be scraped. As the language of all the words in the scarped database cannot be automatically labelled, we are in the process of implementing a human-in-the-loop system \cite{wu2022humaninloop} that can allow human annotation along with automatic labelling approaches.

\section{Analysis of Hansard Māori-English Labelled Database}
\label{section:db_analysis}

An analysis of the language-labelled Hansard database is presented in this section. We present the metadata available, along with a year-wise analysis of the words corresponding to each label. A summary of the foreign words in the corpus is reported, followed by an analysis of the most frequently occurring words in the database. We conclude the analysis with an n-grams analysis and an analysis of the length of sentences in the database.

\subsection{Metadata available}
The Hansard reports available on the New Zealand Parliament website contain the date of the report. We have retained the date of the report in the gold standard Hansard database. Additionally, a number corresponding to the sentence number in the originally scraped document is also included. The metadata available, which are the date of the report and the sentence number, are unique identifiers to every sentence in the Hansard database.

\subsection{Year-wise analysis}
\begin{figure*}[!h]
    \centering
  \includegraphics[scale = 0.45]{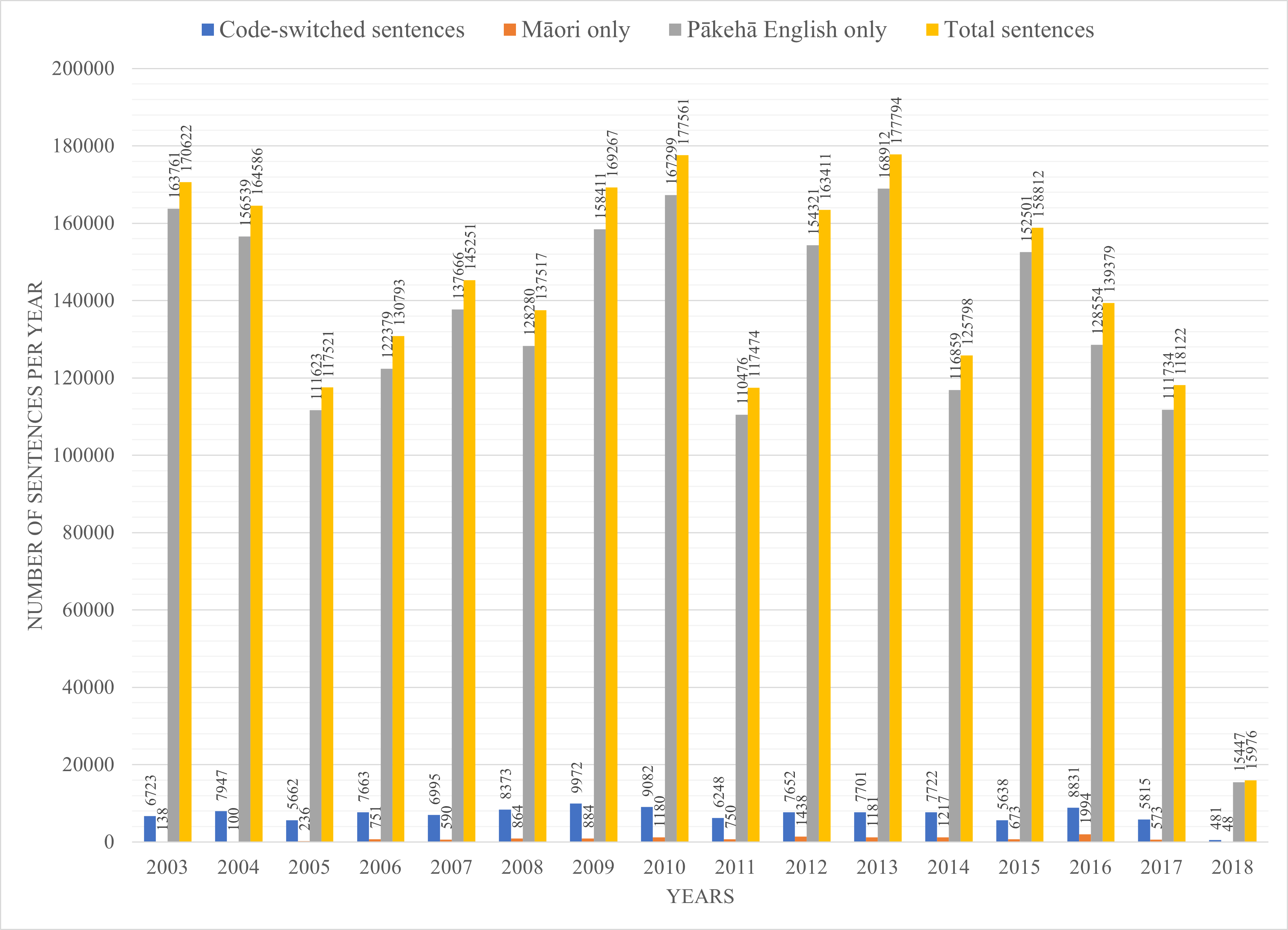}
    \caption{Bar graph showing the number of code-switched, Māori-only, Pākehā English-only and total sentences in the database per year. All available sentences in 2018 were not collected.}
    \label{fig:cs_sentences_count}
\end{figure*}

\begin{figure*}[!h]
    \centering
  \includegraphics[scale = 0.45]{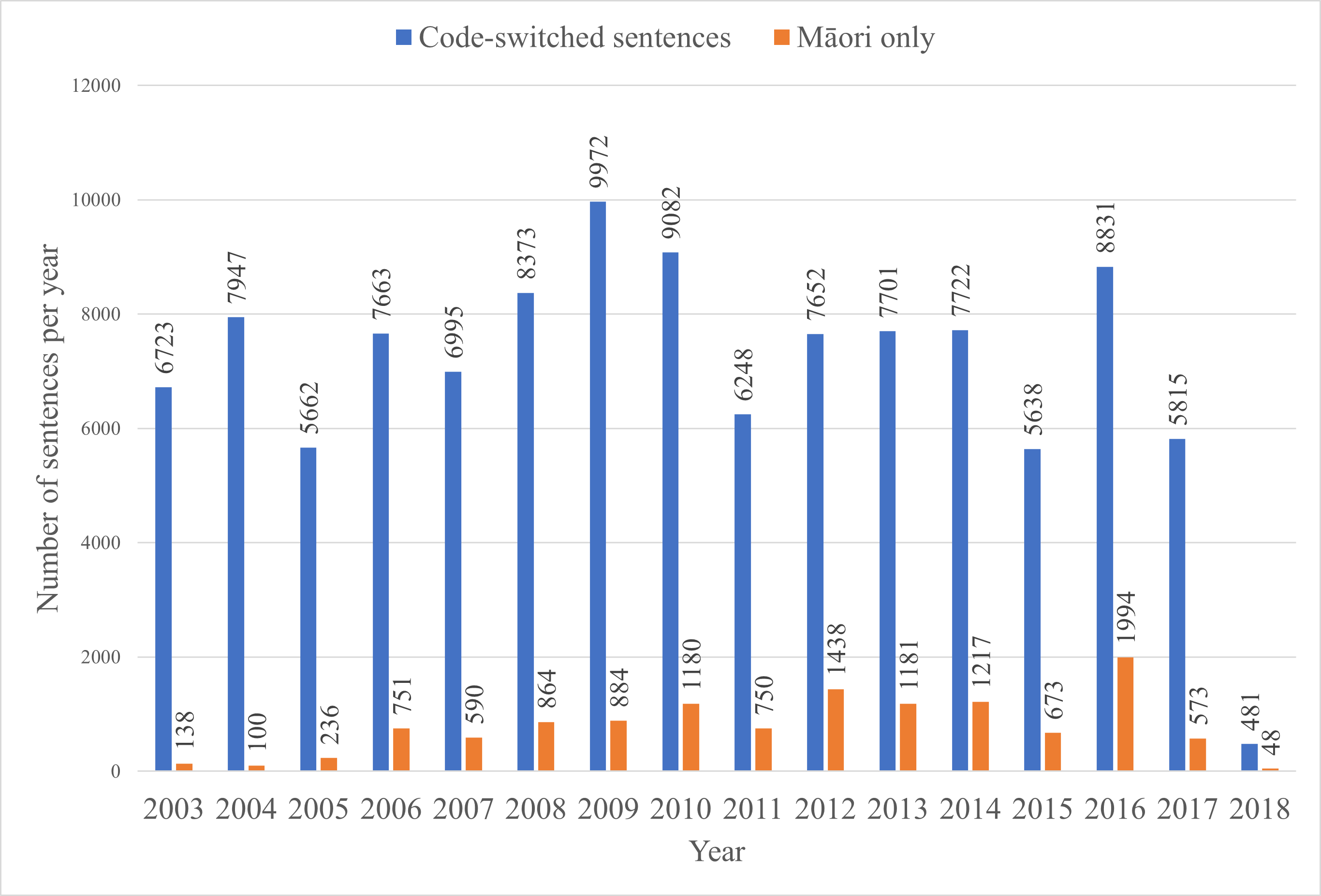}
    \caption{Bar graph showing the number of bilingual and Māori-only sentences in the database per year. All available sentences in 2018 were not collected.}
    \label{fig:cs_sentences_count_only_TRM}
\end{figure*}

This section summarises a year-wise analysis of sentences and words in the database. Figure \ref{fig:cs_sentences_count} provides a bar graph of the number of bilingual, Māori-only, Pākehā English-only and total sentences in the database per year. It can be seen that there are an average of approximately 140,000 sentences per year in the database, excluding 2018. All the available sentences in the Hansard debates webpage in 2018 were not scraped because these webpages followed inconsistent formatting during the time of database collection of this project. This explains the very low number of sentences in 2018. On average, from 2003 to 2018, there were approximately 7500 bilingual sentences, 840 Māori-only sentences and 140,000 Pākehā English-only sentences in the database per year. Figure \ref{fig:cs_sentences_count_only_TRM} provides details of the bilingual and Māori-only sentences in the database per year. It can be seen that the number of bilingual and Māori-only sentences has increased over the years till 2010, and then they have remained at similar numbers. 

\begin{figure*}[!h]
    \centering
  \includegraphics[scale = 0.45]{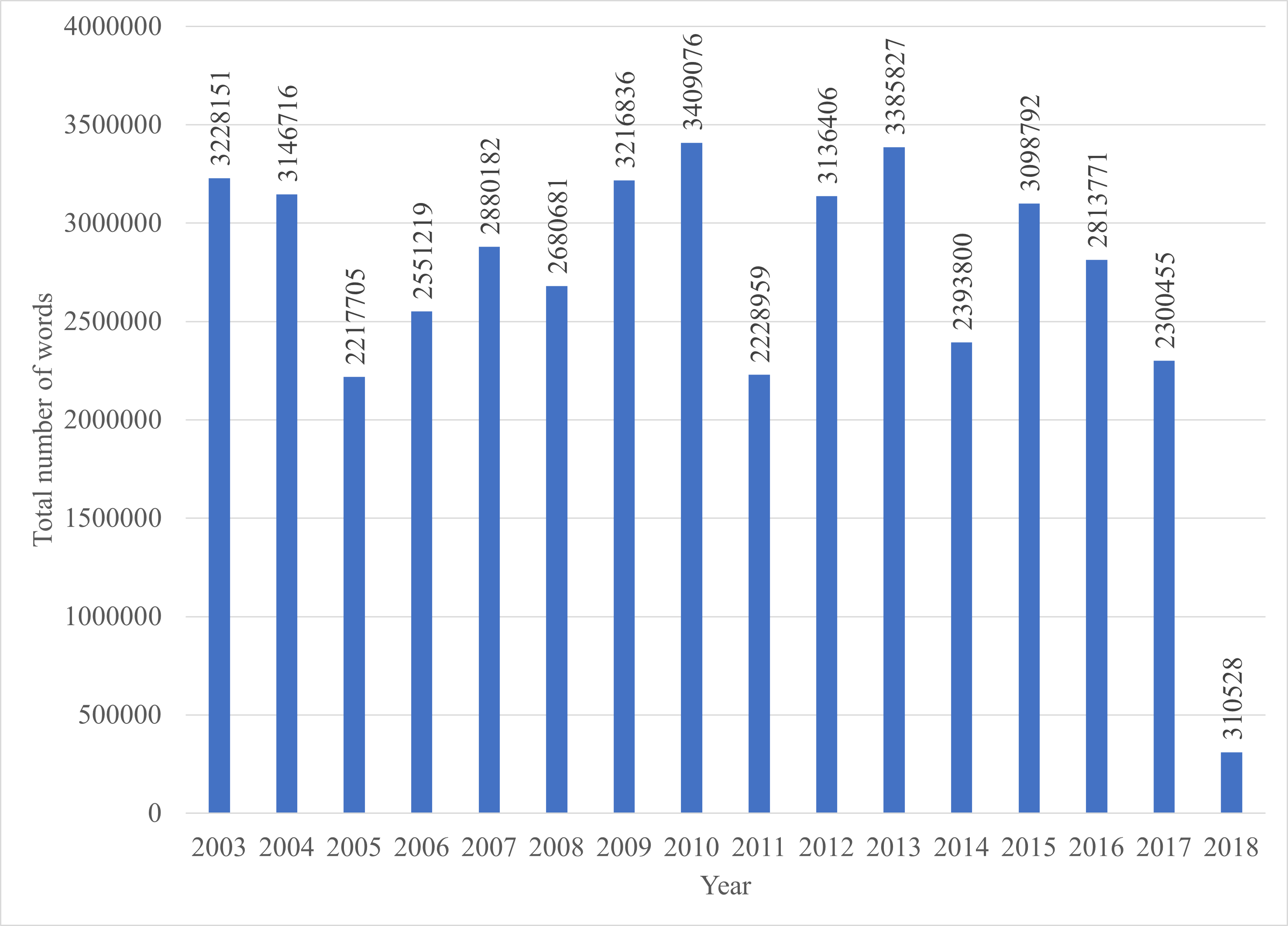}
    \caption{Bar graph showing the number of words in the database per year. All available sentences in 2018 were not collected.}
    \label{fig:cs_worsds_count}
\end{figure*}

\begin{figure*}[!h]
    \centering
  \includegraphics[scale = 0.45]{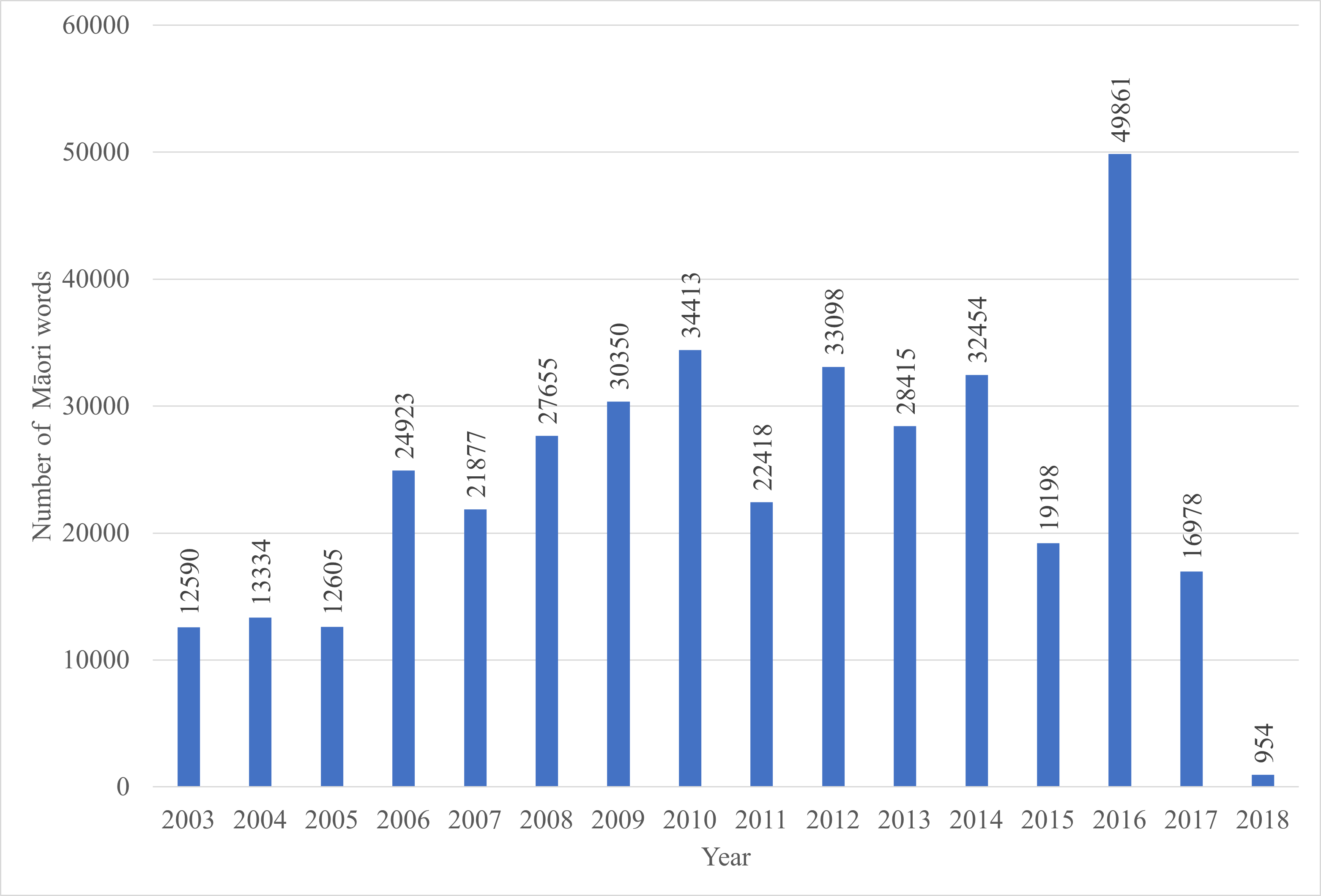}
    \caption{Bar graph showing the number of Māori-only words in the database per year. All available sentences in 2018 were not collected.}
    \label{fig:cs_words_count_only_TRM}
\end{figure*}

Figure \ref{fig:cs_worsds_count} gives a bar graph illustrating the number of words added to the database per year from 2003 to 2018. On average, approximately 2,850,000 words were added to the database annually, with an average of approximately 25,000 Māori words added per year from 2003 to 2017 (based on Figure \ref{fig:cs_words_count_only_TRM}). Again, the numbers from 2018 have not been included in the average calculation due to all data not being collected that year.

\subsection{Foreign words in the database}

\begin{figure*}[!h]
    \centering
  \includegraphics[scale = 0.5]{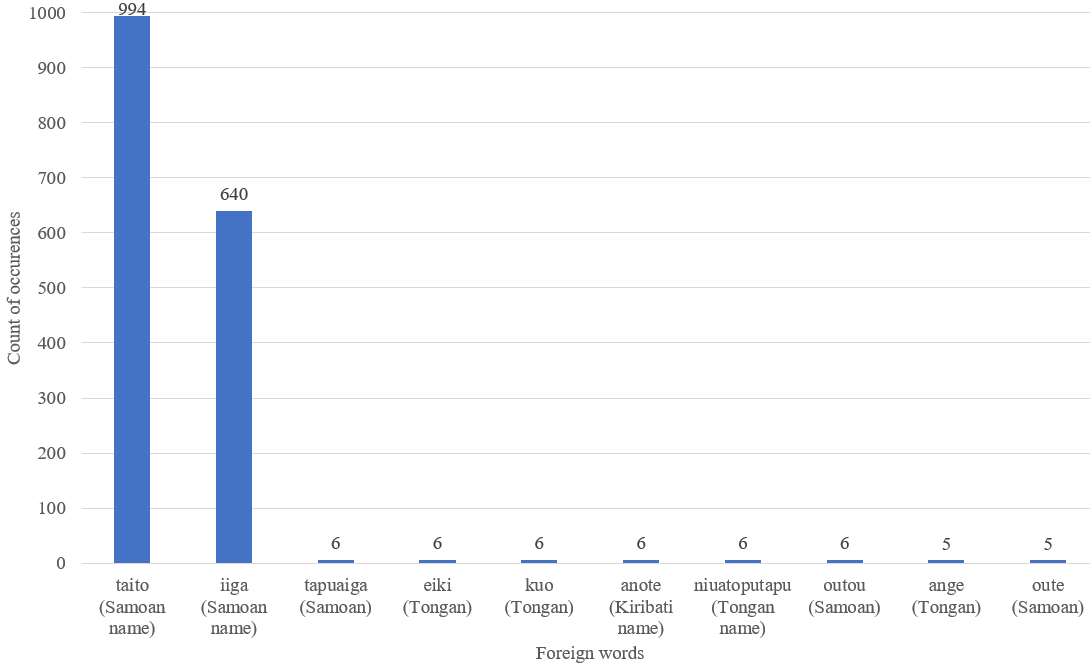}
    \caption{Foreign words in the Hansard database}
    \label{fig:foreignwords}
\end{figure*}

We identified $4025$ foreign words in the Hansard database. Foreign words are words that are neither Pākehā English nor Māori, and do not have English or Māori homographs. Figure \ref{fig:foreignwords} illustrates the $10$ most frequently appearing foreign words in the Hansard database, along with their corresponding language marking. The other languages in the Hansard database include Samoan, Tongan, Kiribati, Cook Islands Māori, Punjabi, Tahitian, Fijian, Bislama, Tuvaluan, Hindi and Hawaiian. There are also Chinese, Japanese, Brazilian, and Papua New Guinean names in the database. The presence of these foreign words is a reflection of the language reality of New Zealand, which has close ties with other Pacific Islands and countries bordering the Pacific Ocean. All the foreign words were marked under the same `F' category. Allocating exact language labels to foreign words is out of the scope of this research. The sentences containing the `F' category words were removed from the final Hansard database, which contains only te reo Māori and English.



\subsection{Words by frequency}

\begin{figure*}
    \centering
  \includegraphics[width=0.49\textwidth,height=6cm]{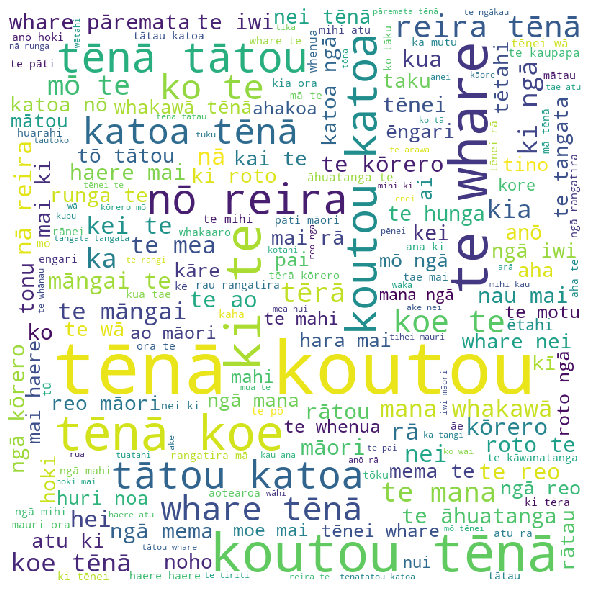}
 \includegraphics[width=0.49\textwidth,height=6cm]{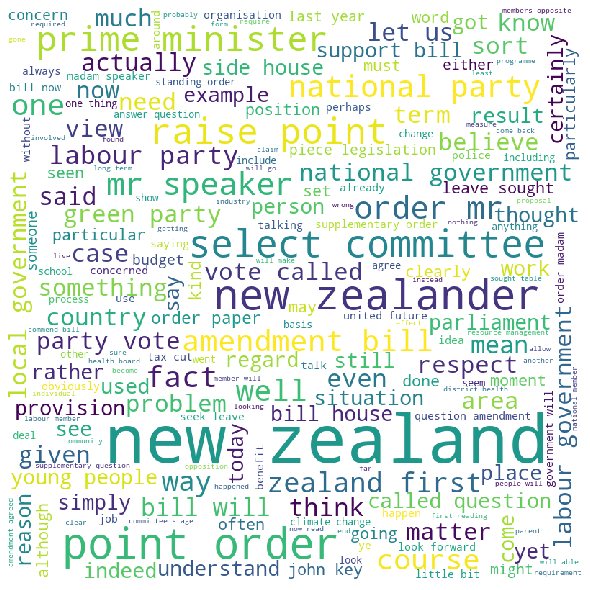}
    \caption{The Hansard Database word cloud: M\={a}ori words (left) and English words (right). }
    \label{fig:wordcloud}
\end{figure*}

This section presents an analysis of the frequently occurring words in the database. Separating the English and Māori words in the database, Figure~\ref{fig:wordcloud} provides a word cloud showing the most common  M\={a}ori words and the most common English words in the Hansard database. The font size of the words is directly proportional to the frequency of occurrences, with the word 'the' occurring $2,809,408$ times and the word  'matter' occurring $24,152$ times. Almost all of the top 200 words by frequency are English, except 'Māori' occurring $38,654$ times and 'te' occurring $28,985$ times. From the Māori word cloud, the most frequent words `tēnā', `koutou', `koe', `katoa', `tātoa' are all part of Māori words for greetings and would be part of the greetings that speakers in the NZ parliament would normally use. The other most common Māori words are:
\begin{itemize}
    \item `whare' meaning house, indicating the many references to parliament house (whare pāramata)
    \item `nō' a particle meaning of, belonging to, from
    \item `reira' meaning the place, time or circumstance mentioned before
    \item `kōrero' meaning to tell, say, speak, read, talk, address\footnote{All meanings were acquired from the \url{https://maoridictionary.co.nz/}.}
\end{itemize}
Analysing the English words word cloud, the most common words are 
\begin{itemize}
    \item New Zealand
    \item New Zealander
    \item Select committee
    \item Point order
\end{itemize}
These words are expected in a New Zealand political context.

\begin{figure*}[!h]
    \centering
  \includegraphics[scale = 0.45]{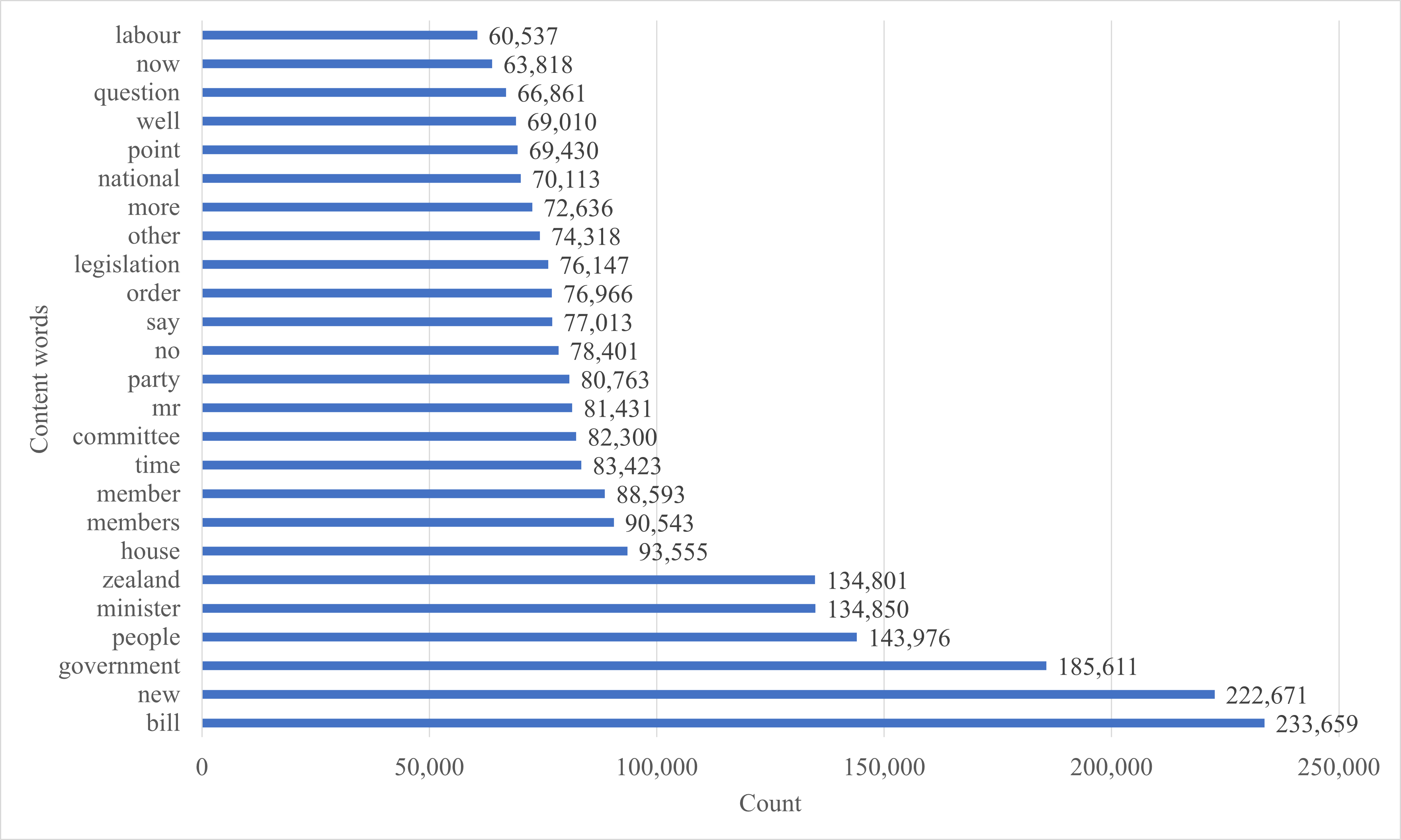}
    \caption{Top 25 most occurring content words from the Hansard database along with the count of the occurrence of each word.}
    \label{fig:25words}
\end{figure*}

\begin{figure*}[!h]
    \centering
  \includegraphics[scale = 0.5]{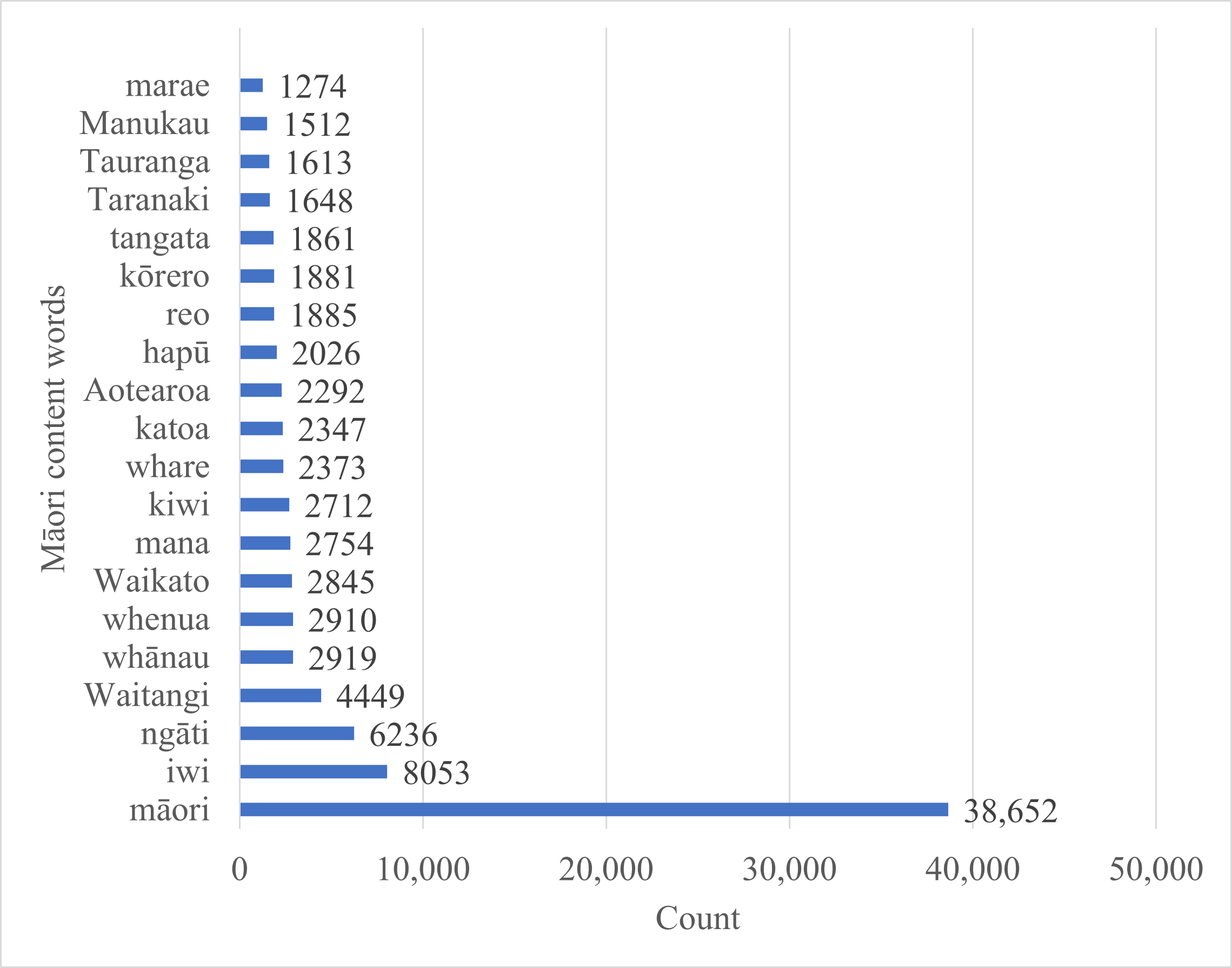}
    \caption{Top 20 most occurring Māori content words from the Hansard database along with the count of the occurrence of each word.}
    \label{fig:20_Maori_words}
\end{figure*}

Considering only the content words, Figure \ref{fig:25words} provides a bar chart of the top 25 content words in the Hansard database. We can notice that all the words are in English. Also, the top 25 words are predominantly political vocabulary.

Further, separating the Māori content words, Figure \ref{fig:20_Maori_words} shows a bar chart of the 20 most common Māori content words in the Hansard database. Here, the words with the first letter capitalised are place names. 

\subsection{N-gram analysis}

\begin{figure}
    \centering
    \includegraphics[width=0.75\textwidth]{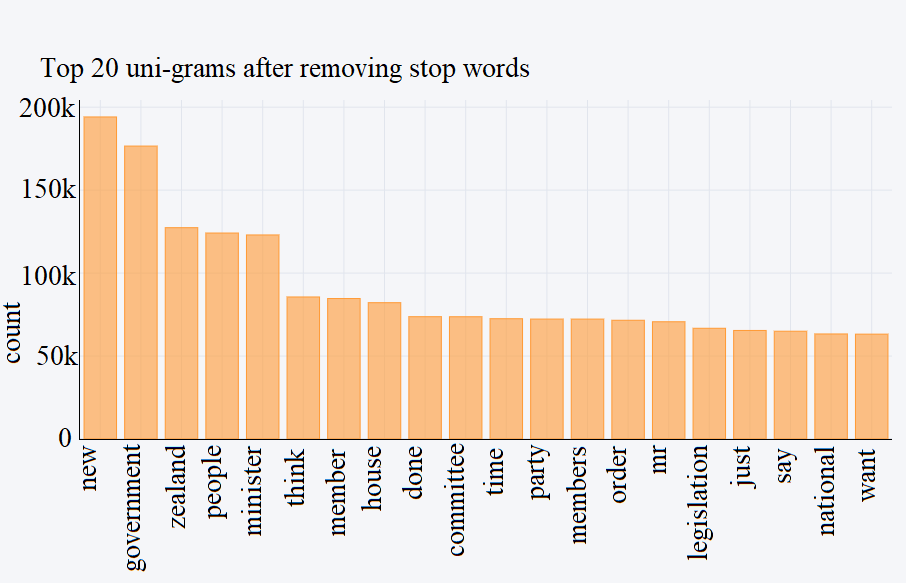}
     \includegraphics[width=0.75\textwidth]{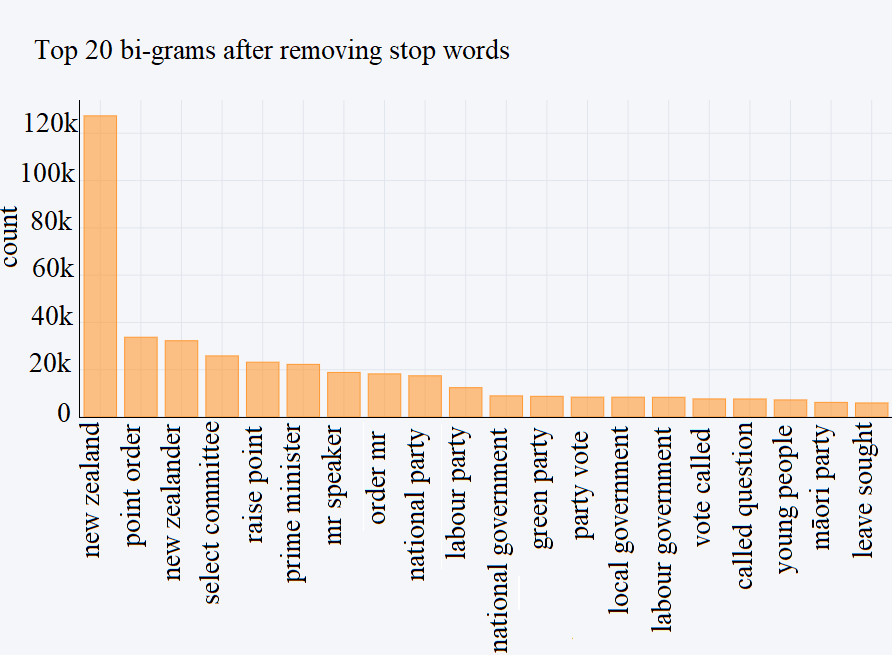}
      \includegraphics[width=0.75\textwidth]{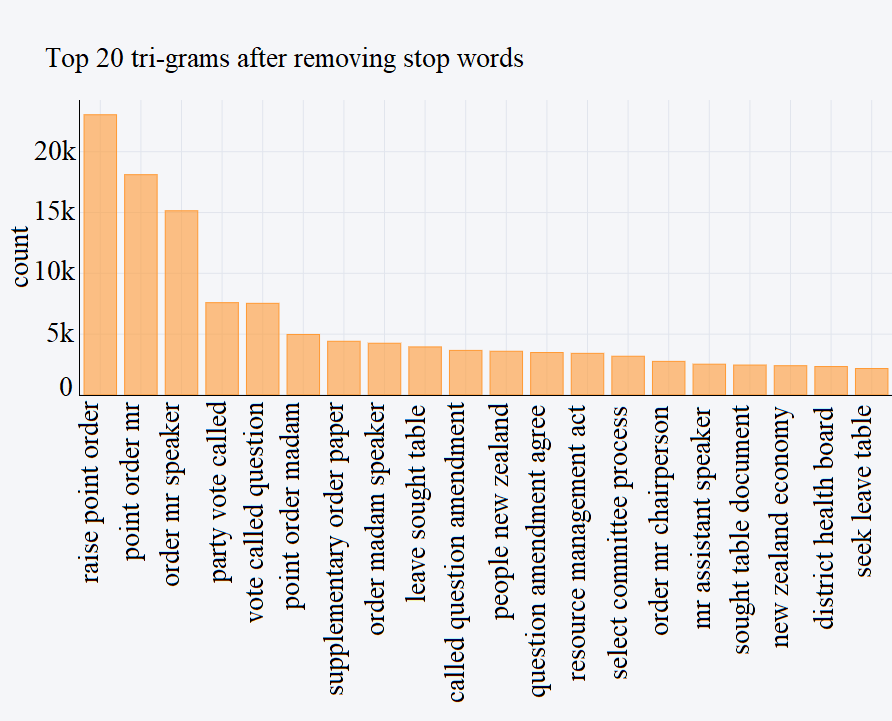}
    \caption{Top 20 uni-grams, bi-grams and tri-grams in the Hansard database} \label{tab:ngram}
\end{figure}

Top uni-grams, bi-grams and tri-grams in the Hansard database are listed in Figure \ref{tab:ngram} along with the number of occurrences of the particular n-gram. Scikit-learn's Count Vectorizer function is used to obtain the list in the table. Only content words were used for this analysis. It can be seen from the table that the political nature of the Hansard database is reflected in the common uni-grams, bi-grams and tri-grams. "Māori" as part of the bi-gram `māori party' is the only Māori word on the list.

\subsection{Text Lengths by Sentence Labels}
\begin{figure}[H]
    \centering
    \includegraphics[width=0.95\textwidth]{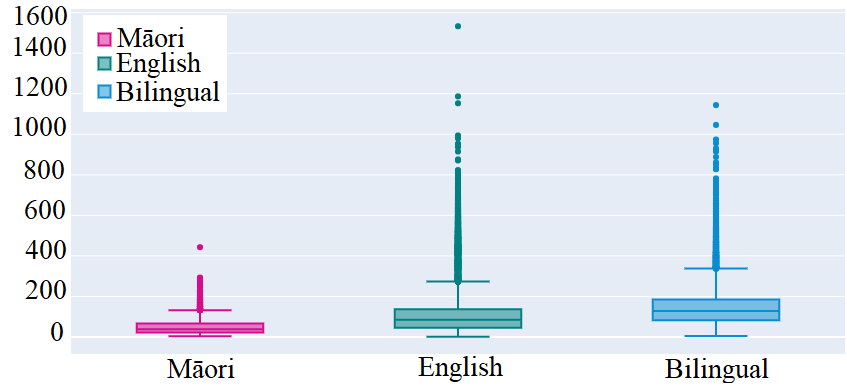}
  \caption{Box plot showing the text lengths by sentence labels in the Hansard database.}
    \label{fig:text_label_boxplot}
\end{figure}

For this analysis, each sentence in the Hansard database was marked as either Māori only, Pākeha (English) only or bilingual based on the word level language labels. The marking was done as:
\begin{itemize}
    \item If all words in the sentence are Māori; then sentence marking = Māori
    \item If all words in the sentence are English; then sentence marking = Pākehā (English)
    \item If there is a mix of Māori and English words in a sentence; then sentence marking = Bilingual
\end{itemize}
Figure \ref{fig:text_label_boxplot} illustrates a box plot of the length of sentences in each of the three categories. We can see that the bilingual sentences are the longest, closely followed by the Pākehā (English) sentences. The Māori sentences are the shortest. There are a large number of outliers for the Pākehā (English) sentences, with the longest sentences having a length of 1500 characters approximately. The long sentences in the database, such as those with a character count of 1000 and more, are generally a list of activities, considerations or names of bills/amendments presented at the Parliament. The Māori sentences have less variation and length, with the longest sentence being approximately 450 characters.  

With this, we conclude the preliminary analysis of the labelled Hansard database. 

\section{Discussion} \label{section:discussion}
The contributions of this study are two-fold. The first one is the development of a labelled bilingual corpus of Māori and English. And, the second contribution is the description of the process of developing such a corpus involving a combination of automatic labelling and hand correction.

The Māori-English labelled corpus is a much-needed resource for language and speech technology development. This corpus is useful for developing language models, language detection, code-switching detection and similar language technology for Māori-English language pairs. A language detection model was trained using the labelled Hansard corpus, and the results are reported in \cite{James_TRM_2022}. The language detection model was tested in a formal context (such as political context in the Hansard database) and informal context (tweets from the \cite{trye2019maori}) with greater than 80\% accuracy. The language detection model was also extended using the labelled Hansard corpus to detect Māori-English code-switching points in the text. This was also tested on the labelled Hansard database, and the Reo Māori Tweets corpus \cite{trye2022harnessing}. The labelled Hansard corpus is currently being used to improve language detection and code-switch detection models.

The process of the labelled corpus development is also presented in detail in this paper. The methodology used for the corpus development can be used by text data in any language with the goal of labelling the languages. The overall process of scraping data, preprocessing and cleaning the data, and categorising the possible labels are all processes that may be common for any language data considered. This can be adopted by other languages as well. However, some steps in the methodology were decided by observing our data and understanding the languages. For example, the decision to apply conditional marking to Ambiguous words (Stage 4 of the Figure \ref{fig:label_block}) was based on observing the Hansard database. Researchers working on other languages and different databases will have to make similar decisions based on their specific contexts.

The linguistic structure of Māori \cite{harlow2007maori, Bauer_maori_1993} and its differences from English have been studied and documented over the years. This knowledge can easily be framed into rules for language classification. This is a quick and less data-intensive approach that can be easily replicated for other low-resource languages that have good documentation of their linguistic structure. Another advantage of a rule-based approach is that it does not require a database to train on, and hence, it is much more suitable for a low-resource setting. Rather than relying on a large amount of language data, we are relying on the years of linguistic knowledge already available regarding language structure. Other language pairs \cite{turkish_english_2018, Kasmuri2020SegregationOC} have also used rule-based approaches as the first or intermediate steps in their code-switching detection systems.

The development of this corpus further solidified the need for researchers familiar with the Māori language (as well as English, of course). The significant hand-labelling tasks and decisions made during the database development required knowledge of te reo Māori. Further, suppose modifications need to be made to the decisions. In that case, the number of categories (Stage 3 of Figure \ref{fig:label_block}), additions to dictionaries and evaluation with the automatic labelling is accurate will all require expert Māori language knowledge. In the future use of the labelled corpus to develop language technology for te reo Māori, expert language knowledge is again required to ensure the best outcomes for Māori speakers.

The hand-correction stages of corpus-building are very time-consuming. We used our existing Māori dictionary to help with this process \cite{James_2020_TRMTTS, shieldste}. But we also acknowledge that our dictionary was not as large as the other ones available (e.g. \url{https://maoridictionary.co.nz}). Perhaps, using larger dictionaries can speed up the hand-correction stages of corpus building. Also, we note that the language detection models being developed using the labelled Hansard corpus \cite{James_TRM_2022} will help future researchers obtain automatically labelled corpora with less time than the one developed here.

The time factor in developing the labelled database was a major bottleneck for not only our research but also for similar research on low-resourced languages \cite{NLP_Time_2021, Crowdsourcing_2012}. A quicker approach was to ignore all sentences containing ambiguous or unclear words. However, with the Māori-English language pair being low-resourced, a loss of 15\% of the available database was not acceptable. Furthermore, ignoring ambiguous words would result in excluding high-frequency words in both Māori and English. For low-resource languages, allocating time for database development, collection, and labelling is inevitable and necessary. The resources developed for these languages benefit technology development in the long run.

We achieved our goal of putting together a labelled bilingual corpus of Māori and English, but the database is also biased. The Hansard database is, as we have demonstrated, predominately composed of English words and is skewed towards political content. We can now effectively analyse code-switching in a `political' context, but the effectiveness of the corpus in an informal context is unknown. Preliminary investigations into detecting code-switching in tweets \cite{James_TRM_2022} have proven successful, but a detailed investigation is necessary. 

Since code-switching appears more in informal contexts such as social media, future research should focus on developing labelled corpora containing social media text. Some attempts in this regard have been reported \cite{trye2019maori, trye2022harnessing}, but more rigorous testing of these corpora in language processing applications is still to be done.

We also acknowledge the difficulty in finding Māori-English bilingual data that is not heavily English-centric. This is a reflection of the language reality of New Zealand, and resources that have a balanced distribution of Māori and English in written text are limited.

While our rule- and dictionary-based approach to language marking successfully reduced the amount of hand-correction necessary, it also introduced error into our database labelling. New Zealand is multicultural and home to speakers of various languages other than Māori and English. These other languages are present in the Hansard database, and some fit the criteria of our rule-based marking or share spelling with Māori or English, meaning they are incorrectly labelled. Although our goal for language detection was restricted to Māori and English, it is important to acknowledge the limitations of the approach and the database.

\section{Conclusions and Future Steps} \label{section:conclusion}

This research focused on developing the first gold-standard Māori-English bilingual database with language labels developed by automatic labelling and manual corrections. While developing this database was time-consuming, it is an invaluable tool for Māori-English language tools research and development. This gold standard is being used to train, test and evaluate rule-based and deep learning models for Māori-English code-switching detection. Several other M\={a}ori, English and bilingual text resources were obtained from various resources, including private sources such as trusted researchers. We share our labelled database (upon request) to enable future growth in this area of research - bound by the Kaitiakitanga License.

However, it is essential to emphasise that there are additional sentences with other languages from the Pacific and abroad. Incorporating these languages and considering it as a multilingual code-switching task is a potential avenue in future.  

The major contribution of this research is the gold-standard database,  thereby developing resources for language and speech technology for New Zealand. The database development recipes and resources developed will be useful for other low-resource languages for similar language technology development.

\section*{Acknowledgements}
The authors thank the researchers who shared their text data with us. We also thank the University of Auckland Faculty of Engineering Research and Development Fund (FRDF) for supporting this project.

\bibliography{biblio}


\begin{thebibliography}{28}
\ifx \bisbn   \undefined \def \bisbn  #1{ISBN #1}\fi
\ifx \binits  \undefined \def \binits#1{#1}\fi
\ifx \bauthor  \undefined \def \bauthor#1{#1}\fi
\ifx \batitle  \undefined \def \batitle#1{#1}\fi
\ifx \bjtitle  \undefined \def \bjtitle#1{#1}\fi
\ifx \bvolume  \undefined \def \bvolume#1{\textbf{#1}}\fi
\ifx \byear  \undefined \def \byear#1{#1}\fi
\ifx \bissue  \undefined \def \bissue#1{#1}\fi
\ifx \bfpage  \undefined \def \bfpage#1{#1}\fi
\ifx \blpage  \undefined \def \blpage #1{#1}\fi
\ifx \burl  \undefined \def \burl#1{\textsf{#1}}\fi
\ifx \doiurl  \undefined \def \doiurl#1{\url{https://doi.org/#1}}\fi
\ifx \betal  \undefined \def \betal{\textit{et al.}}\fi
\ifx \binstitute  \undefined \def \binstitute#1{#1}\fi
\ifx \binstitutionaled  \undefined \def \binstitutionaled#1{#1}\fi
\ifx \bctitle  \undefined \def \bctitle#1{#1}\fi
\ifx \beditor  \undefined \def \beditor#1{#1}\fi
\ifx \bpublisher  \undefined \def \bpublisher#1{#1}\fi
\ifx \bbtitle  \undefined \def \bbtitle#1{#1}\fi
\ifx \bedition  \undefined \def \bedition#1{#1}\fi
\ifx \bseriesno  \undefined \def \bseriesno#1{#1}\fi
\ifx \blocation  \undefined \def \blocation#1{#1}\fi
\ifx \bsertitle  \undefined \def \bsertitle#1{#1}\fi
\ifx \bsnm \undefined \def \bsnm#1{#1}\fi
\ifx \bsuffix \undefined \def \bsuffix#1{#1}\fi
\ifx \bparticle \undefined \def \bparticle#1{#1}\fi
\ifx \barticle \undefined \def \barticle#1{#1}\fi
\bibcommenthead
\ifx \bconfdate \undefined \def \bconfdate #1{#1}\fi
\ifx \botherref \undefined \def \botherref #1{#1}\fi
\ifx \url \undefined \def \url#1{\textsf{#1}}\fi
\ifx \bchapter \undefined \def \bchapter#1{#1}\fi
\ifx \bbook \undefined \def \bbook#1{#1}\fi
\ifx \bcomment \undefined \def \bcomment#1{#1}\fi
\ifx \oauthor \undefined \def \oauthor#1{#1}\fi
\ifx \citeauthoryear \undefined \def \citeauthoryear#1{#1}\fi
\ifx \endbibitem  \undefined \def \endbibitem {}\fi
\ifx \bconflocation  \undefined \def \bconflocation#1{#1}\fi
\ifx \arxivurl  \undefined \def \arxivurl#1{\textsf{#1}}\fi
\csname PreBibitemsHook\endcsname

\bibitem{Eliasson_1989_TRM}
\begin{bchapter}
\bauthor{\bsnm{Eliasson}, \binits{S.}}:
\bctitle{{English-Maori language contact: code-switching and the free-morpheme
  constraint}}.
In: \bbtitle{Reports from Uppsala University Department of Linguistics},
pp. \bfpage{1}--\blpage{28}
(\byear{1989})
\end{bchapter}
\endbibitem

\bibitem{bird-2020-decolonising}
\begin{bchapter}
\bauthor{\bsnm{Bird}, \binits{S.}}:
\bctitle{Decolonising speech and language technology}.
In: \bbtitle{Proceedings of the 28th International Conference on Computational
  Linguistics},
pp. \bfpage{3504}--\blpage{3519}.
\bpublisher{International Committee on Computational Linguistics},
\blocation{Barcelona, Spain (Online)}
(\byear{2020}).
\doiurl{10.18653/v1/2020.coling-main.313}.
\burl{https://aclanthology.org/2020.coling-main.313}
\end{bchapter}
\endbibitem

\bibitem{James_2020_TRMTTS}
\begin{bchapter}
\bauthor{\bsnm{James}, \binits{J.}},
\bauthor{\bsnm{Shields}, \binits{I.}},
\bauthor{\bsnm{Berriman}, \binits{R.}},
\bauthor{\bsnm{Keegan}, \binits{P.J.}},
\bauthor{\bsnm{Watson}, \binits{C.I.}}:
\bctitle{Developing resources for te reo māori text to speech synthesis
  system}.
In: \bbtitle{Sojka P., Kopeček I., Pala K., Horák A. (eds) Text, Speech, and
  Dialogue, Lecture Notes in Computer Science},
p. \bfpage{12284}
(\byear{2020})
\end{bchapter}
\endbibitem

\bibitem{Tehiku_Speech_Recognition}
\begin{botherref}
Te reo {Māori} {Speech Recognition} (under development) by te hiku media
\end{botherref}
\endbibitem

\bibitem{mpai_2017}
\begin{bchapter}
\bauthor{\bsnm{Watson}, \binits{C.}},
\bauthor{\bsnm{Keegan}, \binits{P.}},
\bauthor{\bsnm{Maclagan}, \binits{M.}},
\bauthor{\bsnm{Harlow}, \binits{R.}},
\bauthor{\bsnm{King}, \binits{J.}}:
\bctitle{The motivation and development of {MPAi}, a {Māori Pronunciation
  Aid}}.
In: \bbtitle{Interspeech},
pp. \bfpage{2063}--\blpage{2067}
(\byear{2017})
\end{bchapter}
\endbibitem

\bibitem{gumperz1982discourse}
\begin{bbook}
\bauthor{\bsnm{Gumperz}, \binits{J.J.}}:
\bbtitle{Discourse Strategies}.
\bpublisher{Cambridge University Press},
\blocation{Cambridge, UK}
(\byear{1982})
\end{bbook}
\endbibitem

\bibitem{data_guardian}
\begin{botherref}
\oauthor{\bsnm{Stats}, \binits{N.Z.}}:
{Ngā Tikanga Paihere:} a framework guiding ethical and culturally appropriate
  data use.
Guidelines,
8
(2020)
\end{botherref}
\endbibitem

\bibitem{kaitiakitanga_license}
\begin{botherref}
Kaitiakitanga license.
Whare Kōrero Kaitiakitanga License.
\url{https://xn--wharekrero-v3b.nz/kaitiakitanga/}
\end{botherref}
\endbibitem

\bibitem{Data_sov}
\begin{botherref}
Māori data sovereignty network.
Te Mana Raraunga.
\url{https://www.temanararaunga.maori.nz/}
\end{botherref}
\endbibitem

\bibitem{NZ_stats}
\begin{botherref}
\oauthor{\bsnm{Census}, \binits{N.Z.}}:
2018 census totals by topic – national highlights.
Stats NZ.
\url{https://www.stats.govt.nz/information-releases/2018-census-totals-by-topic-national-highlights-updated}
\end{botherref}
\endbibitem

\bibitem{Bauer_maori_1993}
\begin{botherref}
\oauthor{\bsnm{Bauer}, \binits{W.}},
\oauthor{\bsnm{Parker}, \binits{W.}},
\oauthor{\bsnm{Evans}, \binits{T.K.}}:
Māori.
London: Routledge
(1993)
\end{botherref}
\endbibitem

\bibitem{harlow2007maori}
\begin{bbook}
\bauthor{\bsnm{Harlow}, \binits{R.}}:
\bbtitle{Māori: A Linguistic Introduction}.
\bpublisher{Cambridge University Press},
\blocation{Cambridge, UK}
(\byear{2007})
\end{bbook}
\endbibitem

\bibitem{Biggs_1969}
\begin{bbook}
\bauthor{\bsnm{Biggs}, \binits{B.}}:
\bbtitle{Let’s Learn Māori: a Guide to the Study of the Maori Language}.
\bpublisher{A.H.A.W. Reed},
\blocation{Wellington}
(\byear{1969})
\end{bbook}
\endbibitem

\bibitem{king2011maonze}
\begin{barticle}
\bauthor{\bsnm{King}, \binits{J.}},
\bauthor{\bsnm{Maclagan}, \binits{M.}},
\bauthor{\bsnm{Harlow}, \binits{R.}},
\bauthor{\bsnm{Keegan}, \binits{P.}},
\bauthor{\bsnm{Watson}, \binits{C.}}:
\batitle{The{ MAONZE Corpus}: transcribing and analysing {Māori} speech}.
\bjtitle{New Zealand Studies in Applied Linguistics}
\bvolume{17}(\bissue{1}),
\bfpage{32}--\blpage{48}
(\byear{2011})
\end{barticle}
\endbibitem

\bibitem{keegan2015RMT}
\begin{barticle}
\bauthor{\bsnm{Keegan}, \binits{T.T.}},
\bauthor{\bsnm{Mato}, \binits{P.}},
\bauthor{\bsnm{Ruru}, \binits{S.}}:
\batitle{Using {Twitter} in an indigenous language: An analysis of te reo
  {M{\=a}ori} tweets}.
\bjtitle{AlterNative: An International Journal of Indigenous Peoples}
\bvolume{11}(\bissue{1}),
\bfpage{59}--\blpage{75}
(\byear{2015})
\end{barticle}
\endbibitem

\bibitem{trye2022harnessing}
\begin{botherref}
\oauthor{\bsnm{Trye}, \binits{D.}},
\oauthor{\bsnm{Keegan}, \binits{T.T.}},
\oauthor{\bsnm{Mato}, \binits{P.}},
\oauthor{\bsnm{Apperley}, \binits{M.}}:
{Harnessing Indigenous Tweets}: {The Reo M{\=a}ori Twitter} corpus.
Language resources and evaluation,
1--40
(2022)
\end{botherref}
\endbibitem

\bibitem{trye2019maori}
\begin{bchapter}
\bauthor{\bsnm{Trye}, \binits{D.}},
\bauthor{\bsnm{Calude}, \binits{A.S.}},
\bauthor{\bsnm{Bravo-Marquez}, \binits{F.}},
\bauthor{\bsnm{Keegan}, \binits{T.T.A.G.}}:
\bctitle{M{\=a}ori loanwords: a corpus of {New} {Zealand} {English} tweets}.
In: \bbtitle{{Proc. of the 57th Annual Meeting of the Association for
  Computational Linguistics: Student Research Workshop}},
pp. \bfpage{136}--\blpage{142}
(\byear{2019})
\end{bchapter}
\endbibitem

\bibitem{Niupepa_data}
\begin{botherref}
\oauthor{\bsnm{Niupepa}}:
Māori newspapers - {New Zealand Digital Library}.
Ministry of Education , New Zealand.
\url{http://www.nzdl.org/cgi-bin/library.cgi?a=p&p=about&c=niupepa}
\end{botherref}
\endbibitem

\bibitem{LMC_data}
\begin{botherref}
\oauthor{\bsnm{LMC}}:
Legal {Māori} corpus.
Victoria University of Wellington, New Zealand.
\url{http://nzetc.victoria.ac.nz/tm/scholarly/tei-legalMaoriCorpus.html}
\end{botherref}
\endbibitem

\bibitem{Tehiku_media1}
\begin{botherref}
{Te Hiku Media}.
\url{https://tehiku.nz/te-hiku-tech/papa-reo/}
\end{botherref}
\endbibitem

\bibitem{Tehiku_research}
\begin{botherref}
{Corpus of Te Reo derived from the New Zealand Hansard}.
Te Hiku, Media.
\url{https://github.com/TeHikuMedia/nga-tautohetohe-reo}
\end{botherref}
\endbibitem

\bibitem{wu2022humaninloop}
\begin{botherref}
\oauthor{\bsnm{Wu}, \binits{X.}},
\oauthor{\bsnm{Xiao}, \binits{L.}},
\oauthor{\bsnm{Sun}, \binits{Y.}},
\oauthor{\bsnm{Zhang}, \binits{J.}},
\oauthor{\bsnm{Ma}, \binits{T.}},
\oauthor{\bsnm{He}, \binits{L.}}:
A survey of human-in-the-loop for machine learning.
Future Generation Computer Systems
(2022)
\end{botherref}
\endbibitem

\bibitem{James_TRM_2022}
\begin{bchapter}
\bauthor{\bsnm{James}, \binits{J.}},
\bauthor{\bsnm{Yogarajan}, \binits{V.}},
\bauthor{\bsnm{Shields}, \binits{I.}},
\bauthor{\bsnm{Watson}, \binits{C.}},
\bauthor{\bsnm{Keegan}, \binits{P.}},
\bauthor{\bsnm{Jones}, \binits{P.-L.}},
\bauthor{\bsnm{Mahelona}, \binits{K.}}:
\bctitle{Language models for code-switch detection of te reo {Māori and
  English in a Low-resource Setting}}.
In: \bbtitle{Findings of Conference of the North American Chapter of the
  Association for Computational Linguistics: Human Language Technologies}
(\byear{2022})
\end{bchapter}
\endbibitem

\bibitem{turkish_english_2018}
\begin{bchapter}
\bauthor{\bsnm{Yirmibe{\c{s}}o{\u{g}}lu}, \binits{Z.}},
\bauthor{\bsnm{Eryi{\u{g}}it}, \binits{G.}}:
\bctitle{Detecting code-switching between {T}urkish-{E}nglish language pair}.
In: \bbtitle{{EMNLP} Workshop W-{NUT}: The 4th Workshop on Noisy User-generated
  Text},
pp. \bfpage{110}--\blpage{115}
(\byear{2018})
\end{bchapter}
\endbibitem

\bibitem{Kasmuri2020SegregationOC}
\begin{botherref}
\oauthor{\bsnm{Kasmuri}, \binits{E.}}:
Segregation of code-switching sentences using rule-based technique.
Int. J. Advance Soft Compu. Appl, Vol. 12, No. 1
(2020)
\end{botherref}
\endbibitem

\bibitem{shieldste}
\begin{botherref}
\oauthor{\bsnm{Shields}, \binits{I.}},
\oauthor{\bsnm{Watson}, \binits{C.I.}},
\oauthor{\bsnm{Keegan}, \binits{P.J.}},
\oauthor{\bsnm{Berriman}, \binits{R.}},
\oauthor{\bsnm{James}, \binits{j.} \bsuffix{Jesin}}:
Te reo m{\=a}ori voice for tts
\end{botherref}
\endbibitem

\bibitem{NLP_Time_2021}
\begin{bchapter}
\bauthor{\bsnm{Hedderich}, \binits{M.A.}},
\bauthor{\bsnm{Lange}, \binits{L.}},
\bauthor{\bsnm{Adel}, \binits{H.}},
\bauthor{\bsnm{Jannik}, \binits{S.}},
\bauthor{\bsnm{Klakow}, \binits{D.}}:
\bctitle{{A Survey on Recent Approaches for Natural Language Processing in
  Low-Resource Scenarios}}.
In: \bbtitle{Conference of the North American Chapter of the Association for
  Computational Linguistics: Human Language Technologies},
pp. \bfpage{2545}--\blpage{2568}
(\byear{2021})
\end{bchapter}
\endbibitem

\bibitem{Crowdsourcing_2012}
\begin{bchapter}
\bauthor{\bsnm{Sabou}, \binits{M.}},
\bauthor{\bsnm{Bontcheva}, \binits{K.}},
\bauthor{\bsnm{Scharl}, \binits{A.}}:
\bctitle{Crowdsourcing research opportunities: Lessons from natural language
  processing}.
In: \bbtitle{International Conference on Knowledge Management and Knowledge
  Technologies},
pp. \bfpage{1}--\blpage{8}
(\byear{2012})
\end{bchapter}
\endbibitem

\end{thebibliography}

\newpage
\clearpage

\begin{appendices}

\end{appendices}

\end{document}